\def\year{2022}\relax
\documentclass[letterpaper]{article} 
\usepackage{aaai22}  
\usepackage{times}  
\usepackage{helvet}  
\usepackage{courier}  
\usepackage[hyphens]{url}  
\usepackage{graphicx} 
\usepackage{natbib}  
\usepackage{caption} 
\DeclareCaptionStyle{ruled}{labelfont=normalfont,labelsep=colon,strut=off} 
\usepackage{algorithm}
\usepackage{algorithmic}

\usepackage{pifont}
\usepackage{amsmath}
\usepackage{cleveref}
\usepackage{graphicx}
\usepackage{amsmath}
\DeclareMathOperator*{\argmax}{arg\,max}

\usepackage{amsfonts}
\usepackage{xcolor}
\usepackage{booktabs}
\usepackage{multirow}
\usepackage{xspace}
\usepackage{adjustbox}
\newcommand{\bertmap}{\textsf{BERTMap}\xspace}

\usepackage{newfloat}
\usepackage{listings}
\floatstyle{ruled}
\newfloat{listing}{tb}{lst}{}
\floatname{listing}{Listing}
\title{BERTMap: A BERT-Based Ontology Alignment System}
\author{
    Yuan He\textsuperscript{\rm 1},
    Jiaoyan Chen\textsuperscript{\rm 1},
    Denvar Antonyrajah\textsuperscript{\rm 2},
    Ian Horrocks\textsuperscript{\rm 1}
}
\affiliations{
    \textsuperscript{\rm 1} Department of Computer Science, University of Oxford, UK\\
    \textsuperscript{\rm 2} Samsung Research, UK\\
    \{yuan.he,jiaoyan.chen,ian.horrocks\}@cs.ox.ac.uk, denvar.a@samsung.com

}

\usepackage{bibentry}

\begin{document}

\maketitle

\begin{abstract}
Ontology alignment (a.k.a ontology matching (OM)) plays a critical role in knowledge integration. Owing to the success of machine learning in many domains, it has been applied in OM. However, the existing methods, which often adopt ad-hoc feature engineering or non-contextual word embeddings, have not yet outperformed rule-based systems especially in an unsupervised setting. In this paper, we propose a novel OM system named \bertmap which can support both unsupervised and semi-supervised settings. It first predicts mappings using a classifier based on 
fine-tuning the contextual embedding model BERT on text semantics corpora extracted from ontologies, and then refines the mappings through extension and repair by utilizing the ontology structure and logic. Our evaluation with three alignment tasks on biomedical ontologies demonstrates that \bertmap can often perform better than the leading OM systems LogMap and AML.
\end{abstract}

\section{Introduction} \label{intro}

Ontology alignment (a.k.a. ontology matching (OM)) aims at matching semantically related entities from different ontologies.
A relationship (usually equivalence or subsumption) between two matched entities is known as a mapping.
OM plays an important role in knowledge engineering, as a key technique for ontology integration and quality assurance \cite{ontomatching-sota-n-chal}.
The independent development of ontologies often results in heterogeneous knowledge representations with different categorizations and naming schemes.
For example, the class named ``\textit{muscle layer}'' in the SNOMED Clinical Terms ontology is named ``\textit{muscularis propria}'' in the Foundational Model of Anatomy (FMA) ontology. 
Moreover, real-world ontologies often contain a large number of classes, which not only causes scalability issues, but also makes it harder to distinguish classes of similar names and/or contexts but representing different objects.

Traditional OM solutions typically use lexical matching as their basis and combine it with structural matching and logic-based mapping repair. This has led to several classic systems such as LogMap~\cite{logmap} and AgreementMakerLight (AML)~\cite{amrlight} which still demonstrate state-of-the-art performance on many OM tasks.
However, their lexical matching part only considers texts' surface form such as overlapped sub-strings, and cannot capture the word semantics. 
Recently, machine learning has been proposed as a replacement for lexical and structural matching; for example, DeepAlignment~\cite{kolyvakis-etal-2018-deepalignment} and OntoEmma~\cite{ontoalign-in-bio} utilize word embeddings to represent classes and compute two classes' similarity according to their word vectors' Euclidean distance. Nevertheless, these methods adopt either traditional non-contextual word embedding models such as Word2Vec \cite{word2vec}, which only learns a global (context-free) embedding for each word, or use complex feature engineering which is ad-hoc and relies on a large number of annotated samples for training. In contrast, pre-trained transformer-based language representation models such as BERT \cite{devlin-etal-2019-bert} can learn robust contextual text embeddings, and usually require only moderate training resources for fine-tuning. Although these models  perform well in many Natural Language Processing tasks, they have not yet been sufficiently investigated in OM.

\begin{figure*}[t]
\begin{center}
\includegraphics[width=0.99\textwidth]{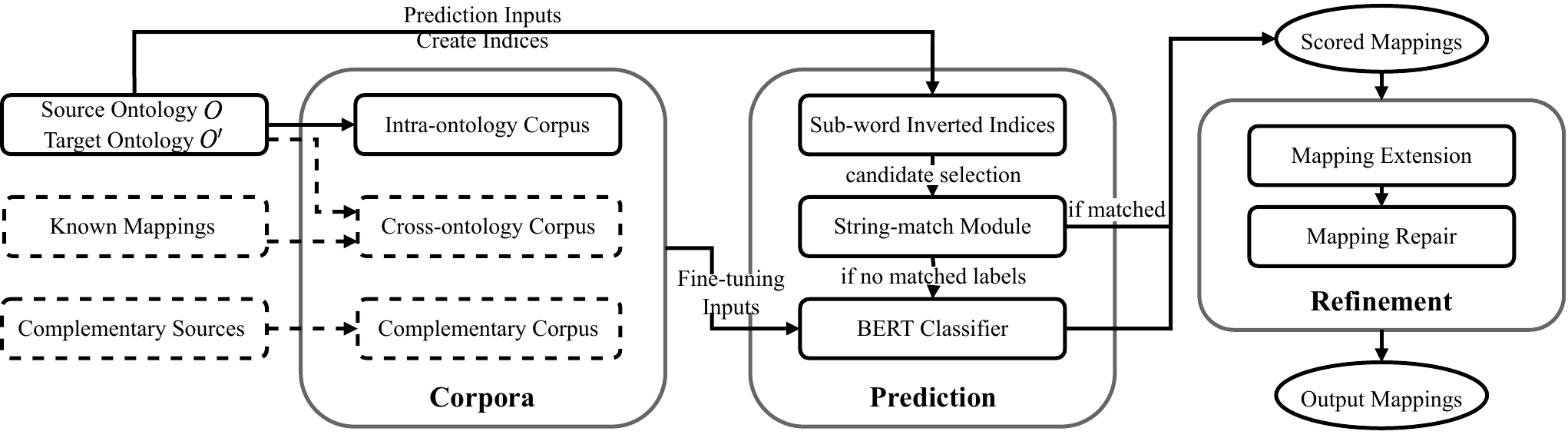}
\caption{Illustration of BERTMap system.}
\label{fig:bertmap}
\end{center}
\end{figure*}

In this paper, we propose \bertmap, a novel ontology alignment system that exploits BERT fine-tuning for mapping prediction and utilizes the graphical and logical information of ontologies for mapping refinement. As shown in Figure \ref{fig:bertmap}, \bertmap includes the following main steps: \textit{(i)} \textit{corpus construction}, where synonym and non-synonym pairs from various sources are extracted; \textit{(ii)} \textit{fine-tuning}, where a suitable pre-trained BERT model is selected and fine-tuned on the corpora constructed in \textit{(i)}; \textit{(iii)} \textit{mapping prediction}, where mapping candidates are first extracted based on sub-word inverted indices and then predicted by the fine-tuned BERT classifier; and \textit{(iv)} \textit{mapping refinement}, where additional mappings are recalled from neighbouring classes of highly scored mappings, and some mappings that lead to logical inconsistency are deleted for higher precision.
 
We evaluate \bertmap\footnote{Codes and data: \url{https://github.com/KRR-Oxford/BERTMap}.} on the FMA-SNOMED task and the FMA-NCI task of the OAEI Large BioMed Track\footnote{\url{http://www.cs.ox.ac.uk/isg/projects/SEALS/oaei/}}, and an extended task of FMA-SNOMED where the more complete labels from the original SNOMED ontology are added.
Our results demonstrate that \bertmap can often outperform the state-of-the-art systems LogMap and AML.

\section{Preliminaries}

\subsection{Problem Formulation} \label{sec:defn}

An ontology is mainly composed of entities (including classes, instances and properties), and axioms that can express relationships between entities.
Ontology alignment involves identifying equivalence, subsumption or other more complex relationships between cross-ontology pairs of entities. In this work, we focus on equivalence between classes.
Given a pair of ontologies, $O$ and $O'$, whose named class sets are $C$ and $C'$, respectively, we aim to first generate a set of scored mappings of the form $(c \in C, c' \in C', P(c \equiv c'))$, where $P(c \equiv c') \in [0, 1]$ is a score indicating the degree to which $c$ and $c'$ are equivalent; we then extend and 
repair the scored mappings to output determined mappings.

\subsection{BERT: Pre-Training and Fine-Tuning} \label{bert:pandf}

BERT is a contextual language representation model 
built on bidirectional transformer encoders 
\cite{attn-is-all}. Its framework involves \textit{pre-training} and \textit{fine-tuning}. In pre-training, the input is a sequence composed of a special token \texttt{[CLS]}, tokens of one sentence $A$, a special token \texttt{[SEP]}, and tokens of another sentence $B$ that follows $A$ in the corpus. Each token's initial embedding encodes its content, its position in the sequence, and the sentence it belongs to ($A$ or $B$). The model has multiple successive layers of an identical architecture. Its main component is the multi-head self-attention block which computes a contextual hidden representation of each token by considering the output of the whole sequence from the previous layer.
The tokens' embeddings from the last layer can be used as the input of a downstream task.
%
Pre-training is conducted by minimizing losses on two tasks: Masked Language Modelling, which predicts a part of tokens that are randomly \textit{masked}, and Next Sentence Prediction, which predicts whether sentence $B$ follows $A$. 
In contrast to the traditional non-contextual word embedding methods 
which assign each token 
only one embedding, BERT distinguishes different occurrences of the same token. For instance, given a sentence \textit{``the bank robber was seen on the river bank''}, BERT computes different embeddings for the two occurrences of \textit{``bank''}, while a non-contextual model yields a unified embedding that is biased towards the most frequent meaning in the corpus. In fine-tuning, pre-trained BERT is attached to customized downstream layers and takes as input either one sentence (e.g., for sentiment classification) or two sentences (e.g., for paraphrasing) according to specific tasks. It typically necessitates only a few epochs and a moderate number of samples for training.


\section{BERTMap} \label{sec:bertmap}


\subsection{Corpus Construction and BERT Fine-Tuning} \label{sec:corpora-fine-tune}

\subsubsection{Text Semantics Corpora}
In real-world ontologies, a named class often has multiple labels (aliases) defined by annotation properties such as \textit{rdfs:label}. 
For convenience, we denote a label after preprocessing\footnote{This includes lowercasing and underscore symbol removing.} by $\omega$, and denote the set of all the preprocessed labels of a class $c$ as $\Omega(c)$. Labels of the same class or from semantically equivalent classes are intuitively synonymous in the domain of the input ontologies; labels from semantically distinct classes can be regarded as non-synonymous. 
The corpora for BERT fine-tuning are composed of pairs of such synonymous labels (i.e., ``\textit{synonyms}'') and pairs of such non-synonymous labels (i.e.,``\textit{non-synonyms}'').
According to the source, the corpora are divided into three categories as follows.

\noindent\textbf{Intra-ontology corpus.} 
For each named class $c$ in an input ontology, we derive all its synonyms which are pairs $(\omega_1, \omega_2)$ with $\omega_1, \omega_2 \in \Omega(c)$, and the special cases where $\omega_1 = \omega_2$ are referred to as \textit{identity synonyms}. 
We consider two types of non-synonyms: \textit{(i) soft non-synonyms} which are labels from two random classes; and  \textit{(ii) hard non-synonyms} which are labels from logically disjoint classes. 
Since class disjointness is often not defined in an ontology, we simply assume that sibling classes (i.e., classes that share a common superclass) are disjoint. 
In fact, this is a naive solution to infer disjointness from the structure of the input ontology.


\noindent\textbf{Cross-ontology corpus.} 
The lack of annotated mappings makes it unfeasible to apply supervised learning on ontology alignment. 
However, it is reasonable to support a semi-supervised setting where a small portion of annotated mappings are given 
and we can extract synonyms from these mappings.
Given a mapping composed of two named classes $c$ and $c'$ we extract all synonyms  $(\omega, \omega')$ where $(\omega, \omega') \in \Omega(c) \times \Omega(c')$ ($\times$ refers to the Cartesian Product).
We also extract non-synonyms from pairs of randomly aligned classes. 

\noindent\textbf{Complementary corpus.} 
We can optionally utilize auxiliary ontologies for additional synonyms and non-synonyms. They are extracted in the same way as the intra-ontology corpus but from an auxiliary ontology.
To reduce data noise and limit the corpus size, we consider auxiliary ontologies of the same domain and only utilize named classes that have shared labels with some class of the input ontologies.

The intra-ontology, cross-ontology and complementary corpora are denoted as $io$, $co$ and $cp$, respectively.
The identity synonyms are denoted as $ids$.
For convenience, we use $+$ to denote the combination of different corpus/synonyms; for example, $io+ids$ refers to the intra-ontology corpus with identity synonyms considered, and $io+co+cp$ refers to including all three corpora without identity synonyms. 
To learn the symmetrical property, we also append reversed synonyms, i.e., if $(\omega_1, \omega_2)$ is in the synonym set, $(\omega_2, \omega_1)$ is added. 
Since some non-synonyms are extracted randomly, they can occasionally also appear in the synonym set; in this case, we delete the non-synonyms.

\subsubsection{Fine-tuning}
Given sets of synonyms and non-synonyms as positive and negative samples, respectively, we fine-tune a pre-trained BERT along with a downstream binary classifier on the cross-entropy loss. Note that we conduct no pre-training but use an existing one from the \texttt{Hugging Face} library\footnote{\url{https://huggingface.co/models}}. 
The inputs of BERT are the tokenized label pairs with the maximum length set to $128$.
The classifier consists of a linear layer (with dropout) that takes as input the embedding of \texttt{[cls]} token from BERT's last-layer outputs, and transforms it into a 2-dimensional vector before applying the output \textit{softmax} layer. The optimization is done using the Adam algorithm \cite{Loshchilov2017FixingWD}. 
The final output is of the form $\langle 1-s, s \rangle$, where $s \in [0,1]$ is the score that indicates the degree that the input label pairs are synonymous.

\subsection{Mapping Prediction} \label{sec:map-pred}

To compute a matched class for each class $c \in C$, a naive solution is to search for $ \argmax_{c' \in C'} P(c \equiv c')$.
Computing mappings in this way has a time complexity of $O(n^2)$, which is impractical for matching large ontologies. To reduce the search space, \bertmap first selects a set of candidates of matched classes using sub-word inverted indices, and then scores each potential mapping with the fine-tuned BERT.

\subsubsection{Candidate Selection 
} \label{sec:sub-ind}

The assumption of our candidate selection is that matched classes are likely to have labels with overlapped sub-tokens. Previous works typically adopt word-level inverted index with additional text processing such as stemming and dictionary consulting \cite{logmap,ontoalign-in-bio}. 
In contrast, \bertmap exploits the sub-word inverted index which can \textit{(i)} capture various word forms without extra processing, and \textit{(ii)} parse unknown words into consecutive known sub-words instead of simply treating them as one special token. 

We build sub-word inverted indices based on BERT's inherent WordPiece tokenizer \cite{wordpiece}, which is trained by an incremental procedure that merges characters (from the corpus) into most likely sub-words at each iteration. 
We opt to use the built-in sub-word tokenizer rather than re-train it on our corpora because it has already been fitted to an enormous corpus (with $3.3$ billion words) that covers various topics \cite{devlin-etal-2019-bert}, and in this context we consider generality to be preferable to task specificity. 

We construct\footnote{Index construction is linear w.r.t.\ the number of sub-words.} indices $I$ and $I'$ for the two input ontologies $O$ and $O'$, respectively.
Each entry of an index is a sub-word, and its values are classes that have at least one label containing this sub-word after tokenization. 
A query of source (resp. target) classes that contain a token $t$ is denoted as $I[t]$ (resp. $I'[t]$). 
The function that takes a class as input and returns all the sub-word tokens of this class's labels is denoted as $T(\cdot)$.
Given a source class $c$, we search from $C'$ the target candidate classes as follows:
we first select target classes that share at least one sub-word token with $c$, i.e., $\bigcup_{t \in T(c)} I'[t]$, and then rank them
according to a scoring metric based on \textit{inverted document frequency (idf)}:
$$
S_{sel}(c, c') = \sum_{t \in T(c) \cap T(c')} idf(t) = \sum_{t \in T(c) \cap T(c')} \log_{10} \frac{|C'|}{|I'[t]|}
$$
where $|\cdot|$ denotes set cardinality. 
Finally, we choose the top $k$ scored target classes for $c$ to form potential mappings of which the scores will be computed.
As a result, we reduce the quadratic time complexity to $O(kn)$ where $k << n$ is the cut-off of candidate selection.

\subsubsection{Mapping Score Computation} \label{sec:map-score}
For a target class candidate $c'$ of the source class $c$, \bertmap uses string matching and the fine-tuned BERT classifier to calculate the mapping score between them as follows:
\begin{equation*}
  S_{map}(c, c') =
    \begin{cases}
      1.0 & \text{if $\Omega(c) \bigcap \Omega(c') \neq \emptyset$}\\
      S_{bert}(\Omega(c), \Omega(c')) & \text{otherwise}
    \end{cases}       
\end{equation*}
where  $\Omega(c) \bigcap \Omega(c') \neq \emptyset$ means $c$ and $c'$ have at least one exactly matched label.
$S_{bert}(\cdot, \cdot)$ 
denotes 
the average of the synonym scores of all the label pairs (i.e., $(\omega, \omega') \in \Omega(c) \times \Omega(c')$), which are predicted by the BERT classifier. The purpose of the string-matching is to save computation by avoiding unnecessary use of the BERT classifier on ``easy'' mappings.
\bertmap finally returns the mapping for $c$ by selecting the top scored candidate $c' = \argmax S_{map}(c, c')$.


With the above steps, we can optionally generate three sets of scored mappings:
\textit{(i)} \texttt{src2tgt} by looking for a matched target class $c' \in C'$ for each source class $c \in C$; \textit{(ii)} \texttt{tgt2src} by looking for a matched source class $c \in C$ for each target class $c' \in C'$; 
and \textit{(iii)} \texttt{combined} by merging \texttt{src2tgt} and \texttt{tgt2src} with duplicates removed. We denote the hyperparameters as $\tau$ and $\lambda$ where $\tau$ refers to the set type (\texttt{src2tgt}, \texttt{tgt2src} or \texttt{combined}) of scored mappings and $\lambda \in [0, 1]$ refers to the mapping score threshold.


\subsection{Mapping Refinement}

\subsubsection{Mapping Extension}
If a source class $c$ and a target class $c'$ are matched, their respective semantically related classes such as parents and children are likely to be matched. This is referred to as the \textit{locality principle} which is assumed in many ontology engineering tasks \cite{Grau2007ALF,JimnezRuiz2020DividingTO}.
\bertmap utilizes this principle to discover new mappings
from those highly scored mappings with  
an \textit{iterative mapping extension} algorithm (see Algorithm \ref{alg:MappingExtension}).
Note that this algorithm only preserves extended mappings that are not previously seen (in $\mathcal{M}$ and $\mathcal{M}_{ex}$) and have scores $\geq \kappa$ (Line $10$ - $12$), i.e., the extension threshold.
Moreover, although $\kappa$ is a hyperparameter, the empirical evidence shows that the results are insensitive to $\kappa$, and thus we set it to a fixed value $\kappa = 0.9$. Finally, the algorithm terminates iteration when no new mappings can be found. 

\begin{algorithm}[tb]
\caption{Iterative Mapping Extension}
\label{alg:MappingExtension}
\textbf{Input}: High confidence mapping set, $\mathcal{M}$\\
\textbf{Parameter}: Extension threshold, $\kappa$\\
\textbf{Output}: Extended mapping set, $\mathcal{M}_{\text{ex}}$\\
\vspace{-0.4cm}
\begin{algorithmic}[1] 
\STATE Initialize the frontier: $\mathcal{M}_{fr} \gets \mathcal{M}$
\STATE Initialize the extended mapping set: $\mathcal{M}_{ex} \gets \{ \}$
\STATE Let $\text{Sup}(\cdot)$ be the function that returns superclasses
\STATE Let $\text{Sub}(\cdot)$ be the function that returns subclasses
\WHILE{$\mathcal{M}_{fr}$ is not empty}
\STATE Initialize an empty new extension set: $\mathcal{M}_{new} \gets \{ \}$
\FOR{each mapping $(c, c', S_{map}(c, c')) \in \mathcal{M}_{fr}$}
\FOR{$(x, x') \in (\text{Sup}(c) \times \text{Sup}(c')) \cup (\text{Sub}(c) \times \text{Sub}(c'))$} 
\STATE $m \gets (x, x', S_{map}(x, x'))$
\IF{$S_{map}(x, x') \geq \kappa$ \AND $m \notin \mathcal{M}$ \AND $m \notin \mathcal{M}_{ex}$}
\STATE $\mathcal{M}_{new} \gets \mathcal{M}_{new} \cup \left\{ m \right\}$ 
\ENDIF
\ENDFOR
\ENDFOR
\STATE $\mathcal{M}_{ex} \gets \mathcal{M}_{ex} \cup \mathcal{M}_{new}$ 
\STATE $\mathcal{M}_{fr} \gets \mathcal{M}_{new}$ 
\ENDWHILE
\STATE \textbf{return} $\mathcal{M}_{ex}$
\end{algorithmic}
\end{algorithm}

\subsubsection{Mapping Repair} 
Mapping repair removes mappings that will lead to logical conflicts after integrating two ontologies. A ``perfect repair'' (a.k.a.\ a \textit{diagnosis}) refers to removing a minimal number of mappings to achieve logical coherence. However, computing a diagnosis is usually time-consuming, and there may be no unique solution. To address this, \citet{JimnezRuiz2013EvaluatingMR} proposes a propositional logic-based repair method that can efficiently compute an \textit{approximate} repair $R$ which ensures that: \textit{(i)} $R$ is a subset of the diagnosis (so that there is no sacrifice of correct mappings); \textit{(ii)} only a small number of unsatisfiable classes remain. Mapping repair is commonly used in classic OM systems, but rarely considered in machine learning-based approaches. In this work, we adopt the repair tool developed by \citet{JimnezRuiz2013EvaluatingMR}. 

Note that mapping extension and repair can consistently improve the performance without excessive time cost, because the former only needs to handle mappings of high prediction scores and the later adopts an efficient repair algorithm \cite{JimnezRuiz2013EvaluatingMR}.

\section{Evaluation} \label{sec:exp}
\begin{table}[t!]
    \centering
    \begin{adjustbox}{width=0.45\textwidth}
    \begin{tabular}{l c c c c}
    \toprule
        \textbf{Task} & \textbf{SRC} & \textbf{TGT} & \textbf{Refs} ($=$) & \textbf{Refs} ($?$) \\\midrule
        FMA-SNOMED & 10,157 & 13,412 & 6,026 & 2,982\\
        FMA-NCI & 3,696 & 6,488 & 2,686 & 338 \\
    \bottomrule
    \end{tabular}
    \end{adjustbox}
    \vspace{-0.2cm}
    \caption{
    Numbers of classes and reference mappings
    in the FMA-SNOMED and FMA-NCI tasks.
    }
    \label{tab:data}
\end{table}
\subsection{Experiment Settings}
\subsubsection{Datasets and Tasks} 
The evaluation considers the FMA-SNOMED and FMA-NCI small fragment tasks of the OAEI LargeBio Track. They have large-scale ontologies and high quality gold standards created by domain experts. Table \ref{tab:data} summarizes the numbers of classes in source (SRC) and target (TGT) ontologies, and the numbers of reference mappings.
``Refs (=)'' refers to the reference mappings to be considered, while ``Refs (?)'' refers to the reference mappings that will cause logical inconsistency after alignment and are ignored as suggested by OAEI.
We also consider an extended task of FMA-SNOMED, denoted as FMA-SNOMED+, where the target ontology is extended by introducing the labels from the latest version of SNOMED.\footnote{The version of 20210131 is available at: \url{https://www.nlm.nih.gov/healthit/snomedct/index.html}.} This is because the LargeBio SNOMED is many years out of date, and the naming scheme in the newly released SNOMED has changed and many more class labels have been added. 
We adopt the following strategy to construct SNOMED+: for each class $c$ in SNOMED, we extract its labels $\Omega(c)$ and for each label $\omega$ in $\Omega(c)$, we 
search for classes in the original SNOMED that have $\omega$ as an alias; we then add all the labels of the searched classes to the LargeBio SNOMED for SNOMED+. 
We also use these additional labels
to construct the complementary corpus for the FMA-SNOMED task.
The key difference is that 
they are used for fine-tuning alone on the FMA-SNOMED task but for both fine-tuning and prediction on the FMA-SNOMED+ task.

\subsubsection{Evaluation Metrics} 
We evaluate all the systems on Precision (P), Recall (R), and Macro-F1 (F1), defined as: 
$$P = \frac{|\mathcal{M}_{out} \cap \mathcal{M}_{=} \backslash \mathcal{M}_{?}|}{|\mathcal{M}_{out} \backslash \mathcal{M}_{?}|}, R = \frac{|\mathcal{M}_{out} \cap \mathcal{M}_{=} \backslash \mathcal{M}_{?}|}{|\mathcal{M}_{=} \backslash \mathcal{M}_{?}|}$$
and $F1 = 2PR / (P + R)$, where $\mathcal{M}_{out}$ is the system's output mappings, $\mathcal{M}_{=}$ and $\mathcal{M}_{?}$ refer to reference mappings to be considered (Refs (=)) and ignored (Refs (?)), respectively. 
In the unsupervised setting, we divide $\mathcal{M}_{=}$ into $\mathcal{M}_{val}$ ($10\%$) and $\mathcal{M}_{test}$ ($90\%$); and in the semi-supervised setting, we divide $\mathcal{M}_{=}$ into $\mathcal{M}_{train}$ ($20\%$), $\mathcal{M}_{val}$ ($10\%$) and $\mathcal{M}_{test}$ ($70\%$). 
When computing the metrics on the hold-out validation or test set, we should regard reference mappings that are not in this set as neither positive nor negative (i.e., as ignored mappings). 
For example, during validation, we add the mappings from  $\mathcal{M}_{train}$ (if semi-supervised) and $\mathcal{M}_{test}$ (for both settings) into $\mathcal{M}_{?}$ when calculating the metrics.

\subsubsection{BERTMap Settings}
We set up various \bertmap settings considering
\textit{(i)} being unsupervised (without $co$) or semi-supervised ($+co$), \textit{(ii)} including the identity synonyms ($+ids$), \textit{(iii)} being augmented with a complementary corpus ($+cp$), and \textit{(iv)} applying mapping extension ($ex$) and repair ($rp$). 
In fine-tuning, the semi-supervised setting takes all the label pairs extracted from both within the input ontologies and $\mathcal{M}_{train}$ as training data, label pairs from $\mathcal{M}_{val}$ as validation data and label pairs from $\mathcal{M}_{test}$ as test data, while the unsupervised setting
partitions all the label pairs extracted from within the input ontologies into $80\%$ for training and $20\%$ for validation. 
Note that the \textit{the validation in fine-tuning} is different from the \textit{mapping validation} which uses $\mathcal{M}_{val}$ because the former concerns the performance of the BERT classifier while the latter concerns selecting the best hyperparameters for determining output mappings. 

Besides, we set the positive-negative sample ratio to $1:4$. Namely, we sample $4$ non-synonyms for each synonym in $co$, and $2$ soft and $2$ hard non-synonyms for each synonym in other corpora. We use Bio-Clinical BERT, which has been pre-trained on biomedical and clinical domain corpora \cite{alsentzer-etal-2019-publicly}. The BERT model is  fine-tuned for $3$ epochs with a batch size of $32$, and evaluated on the validation set for every $0.1$ epoch, through which the best checkpoint on the cross-entropy loss is selected for prediction.
The cut-off of sub-word inverted index-based candidate selection is set to $200$. Our implementation uses \textit{(i)} \texttt{owlready2}\footnote{\url{https://owlready2.readthedocs.io/en/latest/}.} for ontology processing and \textit{(ii)} \texttt{transformers}\footnote{\url{https://huggingface.co/transformers/}.} for BERT. 
The training uses 
a single GTX 1080Ti GPU.

After fine-tuning, we perform a 2-step mapping validation using $\mathcal{M}_{val}$ as follows: we first validate the scored mappings from prediction 
and obtain the best $\{\tau, \lambda\}$; we then extend the mappings by Algorithm \ref{alg:MappingExtension} and 
%
 validate the extended mappings and obtain another best mapping filtering threshold $\lambda$. Interestingly, in all our \bertmap experiment settings, we find the best $\lambda$ obtained in the first step always coincides with the best $\lambda$ obtained in the second step. This demonstrates the robustness of our mapping extension algorithm. After validation, we repair and ouput the mappings. 
Note that we also test \bertmap without extension and repair, and in this case, we skip the second mapping validation step and output mappings with scores $\geq \lambda$.

\subsubsection{Baselines} 
We compare \bertmap with various baselines as follows: \textit{(i)} \textit{String-matching} as defined in the Mapping Score Computation; \textit{(ii)} \textit{Edit-similarity}, which computes the maximum normalized edit similarity between the labels of two classes as their mapping score (note that \textit{(i)} is a special case of \textit{(ii)}); \textit{(iii)} \textit{LogMap} and \textit{AML}, which are the leading systems in many OAEI tracks and other tasks; \textit{(iv)} \textit{LogMapLt}, the lexical matching part of LogMap; \textit{(v)} \textit{LogMap-ML$^{*}$}, which is a variant of LogMap-ML \cite{chen2021augmenting} using no branch conflicts but only LogMap anchor mappings for extracting samples for training, where Word2Vec is used to embed the class label and a Siamese Neural Network with Multilayer Perception is used as the classifier. Note that \textit{(i)} and \textit{(ii)} are our internal baselines, and we set up the same candidate selection and hyperparameter search procedure for them as for \bertmap; whereas \textit{(iii)} to \textit{(v)} are external systems with default implementations.
Note that by comparing to LogMap and AML, we actually have several indirect baselines that have participated in the LargeBio Track (e.g., ALOD2Vec \cite{Portisch2018ALOD2VecM} and Wiktionary \cite{Portisch2019WiktionaryM}).

\subsection{Results}


\begin{table*}[!ht]
    \fontsize{9pt}{9.5pt}
    \selectfont
	\centering
	\begin{tabular}{l c c c c c c c c}
		\toprule
		 & & \multicolumn{3}{c}{90\% Test Mappings} & \phantom{a} & \multicolumn{3}{c}{70\% Test Mappings} \\ 
		 \cmidrule{3-5} \cmidrule{7-9}
		 System & $\{\tau, \lambda\}$ & Precision & Recall & Macro-F1 & & Precision & Recall & Macro-F1 \\ \midrule
		 io & (tgt2src, 0.999) & 0.705 & 0.240 & 0.359 && 0.649 & 0.239 & 0.350 \\ 
		 io+ids & (tgt2src, 0.999) & 0.835 & 0.347 & 0.490 && 0.797 & 0.346 & 0.483 \\ 
		 io+cp & (src2tgt, 0.999) & 0.917 & 0.750 & 0.825 && 0.895 & 0.748 & 0.815 \\ 
		 io+ids+cp & (src2tgt, 0.999) & 0.910 & 0.758 & 0.827 && 0.887 & 0.755 & 0.816 \\ 
		 io+ids+cp (ex) & (src2tgt, 0.999) & 0.896 & 0.771 & 0.829 && 0.869 & 0.771 & 0.817 \\ 
		 io+ids+cp (ex+rp) & (src2tgt, 0.999) & 0.905 & 0.771 & \textbf{0.833} && 0.881 & 0.771 & 0.822 \\ \midrule
		 io+co & (src2tgt, 0.997) & NA & NA & NA && 0.937 & 0.564 & 0.704 \\ 
		 io+co+ids & (src2tgt, 0.999) & NA & NA & NA && 0.850 & 0.714 & 0.776 \\ 
		 io+co+cp & (src2tgt, 0.999) & NA & NA & NA && 0.880 & 0.779 & 0.826 \\ 
		 io+co+ids+cp & (src2tgt, 0.999) & NA & NA & NA && 0.899 & 0.774 & 0.832 \\ 
		 io+co+ids+cp (ex) & (src2tgt, 0.999) & NA & NA & NA && 0.882 & 0.787 & 0.832 \\ 
		 io+co+ids+cp (ex+rp) & (src2tgt, 0.999) & NA & NA & NA && 0.892 & 0.786 & \textbf{0.836} \\ \midrule
		 string-match & (combined, 1.000) & 0.987 & 0.194 & 0.324 && 0.983 & 0.192 & 0.321 \\ 
		 edit-similarity & (combined, 0.920) & 0.971 & 0.209 & 0.343 && 0.963 & 0.208 & 0.343 \\ 
		 LogMapLt & NA & 0.965 & 0.206 & 0.339 && 0.956 & 0.204 & 0.336 \\ 
		 LogMap & NA & 0.935 & 0.685 & 0.791 && 0.918 & 0.681 & 0.782 \\ 
		 AML & NA & 0.892 & 0.757 & 0.819 && 0.865 & 0.754 & 0.806 \\ 
		 $\text{LogMap-ML}^{*}$ & NA & 0.944  & 0.205 & 0.337 && 0.928 & 0.208 & 0.340 \\ 
		 \bottomrule \\
	\end{tabular}
	\vspace{-0.5cm}
	\caption{Results of \bertmap under different settings and baselines on the FMA-SNOMED task.}
	\label{fma2snomed:results}
	\vspace{0.3cm}

	\begin{tabular}{l c c c c c c c c}
		\toprule
		 & & \multicolumn{3}{c}{90\% Test Mappings} & \phantom{a} & \multicolumn{3}{c}{70\% Test Mappings} \\ 
		 \cmidrule{3-5} \cmidrule{7-9}
		 System & $\{\tau, \lambda\}$ & Precision & Recall & Macro-F1 & & Precision & Recall & Macro-F1 \\ \midrule
		 io & (src2tgt, 0.999) & 0.930 & 0.836 & 0.880 && 0.911 & 0.834 & 0.871 \\ 
		 io+ids & (src2tgt, 0.999) & 0.926 & 0.834 & 0.878 && 0.906 & 0.832 & 0.868 \\ 
		 io+ids (ex) & (src2tgt, 0.999) & 0.916 & 0.852 & 0.883 && 0.894 & 0.851 & 0.872 \\ 
		 io+ids (ex+rp) & (src2tgt, 0.999) & 0.924 & 0.851 & \textbf{0.886} && 0.905 & 0.851 & 0.877 \\ \midrule
		 io+co & (src2tgt, 0.999) & NA & NA & NA && 0.913 & 0.841 & 0.875 \\ 
		 io+co+ids & (src2tgt, 0.999) & NA & NA & NA && 0.913 & 0.836 & 0.873 \\ 
		 io+co+ids (ex) & (src2tgt, 0.999) & NA & NA & NA && 0.899 & 0.852 & 0.875 \\ 
		 io+co+ids (ex+rp) & (src2tgt, 0.999) & NA & NA & NA && 0.908 & 0.852 & \textbf{0.879} \\ \midrule
		 string-match & (src2tgt, 1.000) & 0.978 & 0.672 & 0.797 && 0.972 & 0.665 & 0.790 \\ 
		 edit-similarity & (src2tgt, 0.930) & 0.978 & 0.728 & 0.834 && 0.972 & 0.724 & 0.830 \\ 
		 LogMapLt & NA & 0.953 & 0.717 & 0.819 && 0.940 & 0.709 & 0.808 \\ 
		 LogMap & NA & 0.869 & 0.867 & 0.868 && 0.838 & 0.868 & 0.852 \\ 
		 AML & NA & 0.895 & 0.829 & 0.861 && 0.868 & 0.825 & 0.846 \\ 
		 $\text{LogMap-ML}^{*}$ & NA & 0.955  & 0.684 & 0.797 && 0.942 & 0.700 & 0.803 \\ 
		 \bottomrule \\ 
	\end{tabular}
	\vspace{-0.5cm}
	\caption{Results of \bertmap under different settings and baselines on the FMA-SNOMED+ task.}
	\label{fma2snomed+:results}
	\vspace{0.3cm}
	
	\begin{tabular}{l c c c c c c c c}
		\toprule
		 & & \multicolumn{3}{c}{90\% Test Mappings} & \phantom{a} & \multicolumn{3}{c}{70\% Test Mappings} \\ 
		 \cmidrule{3-5} \cmidrule{7-9}
		 System & $\{\tau, \lambda\}$ & Precision & Recall & Macro-F1 & & Precision & Recall & Macro-F1 \\ \midrule
		 io & (src2tgt, 0.999) & 0.930 & 0.847 & 0.887 && 0.912 & 0.851 & 0.880 \\ 
		 io+ids & (src2tgt, 0.999) & 0.936 & 0.842 & 0.887 && 0.920 & 0.845 & 0.881 \\ 
		 io+ids (ex) & (src2tgt, 0.999) & 0.926 & 0.852 & 0.888 && 0.907 & 0.854 & 0.880 \\ 
		 io+ids (ex+rp) & (src2tgt, 0.999) & 0.938 & 0.852 & 0.893 && 0.922 & 0.854 & 0.887 \\ \midrule
		 io+co & (src2tgt, 0.999) & NA & NA & NA && 0.939 & 0.838 & 0.886 \\ 
		 io+co+ids & (src2tgt, 0.999) & NA & NA & NA && 0.961 & 0.805 & 0.876 \\ 
		 io+co+ids (ex) & (src2tgt, 0.999) & NA & NA & NA && 0.955 & 0.813 & 0.879 \\ 
		 io+co+ids (ex+rp) & (src2tgt, 0.999) & NA & NA & NA && 0.959 & 0.813 & 0.880 \\ \midrule
		 string-match & (tgt2src, 1.000) & 0.978 & 0.742 & 0.843 && 0.972 & 0.747 & 0.845 \\ 
		 edit-similarity & (src2tgt, 0.900) & 0.976 & 0.768 & 0.860 && 0.970 & 0.774 & 0.861 \\ 
		 LogMapLt & NA & 0.963 & 0.815 & 0.883 && 0.953 & 0.812 & 0.877 \\ 
		 LogMap & NA & 0.938 & 0.900 & \textbf{0.919} && 0.922 & 0.897 & \textbf{0.909} \\ 
		 AML & NA & 0.936 & 0.900 & 0.918 && 0.919 & 0.898 & \textbf{0.909} \\ 
		 $\text{LogMap-ML}^{*}$ & NA & 0.968 & 0.715 & 0.822 && 0.959 & 0.714 & 0.818 \\ 
		 \bottomrule \\ 
	\end{tabular}
	\vspace{-0.5cm}
	\caption{Results of \bertmap systems under different settings and baselines on the FMA-NCI task.}
	\label{fma2nci:results}
\end{table*}


The results together with the corresponding hyperparameter settings are shown in Tables \ref{fma2snomed:results}, \ref{fma2snomed+:results} and \ref{fma2nci:results}, where 
$90\%$ (resp. $70\%$) Test Mappings refer to the results measured on $\mathcal{M}_{test}$ of the unsupervised (resp. semi-supervised) setting.
To fairly compare the unsupervised and semi-supervised settings, we report the results on both $90\%$ and $70\%$ Test Mappings for the unsupervised setting.

The overall results show that \bertmap can achieve higher F1 score than all the baselines on the FMA-SNOMED and FMA-SNOMED+ tasks, but its F1 score is lower than LogMap and AML on the FMA-NCI task. 
On the FMA-SNOMED task, the unsupervised \bertmap can surpass AML (resp. LogMap) by $1.4\%$ (resp. $4.2\%$) in F1, while the semi-supervised \bertmap can exceed AML (resp. LogMap) by $3.0\%$ (resp. $5.4\%$). 
The corresponding rates become $2.5\%$ (resp. $1.8\%$) and $3.3\%$ (resp. $2.7\%$) on the FMA-SNOMED+ task. 
On the FMA-NCI task, the best F1 score of the unsupervised \bertmap is worse than AML (resp. LogMap) by $2.5\%$ (resp. $2.6\%$), and the best F1 score of the semi-supervised \bertmap is worse than AML (resp. LogMap) by $2.3\%$ (resp. $2.3\%$). 
Note that \bertmap without $ex$ or $rp$ consistently outperforms LogMapLt on all the tasks. This suggests that with a more suitable mapping refinement strategy, \bertmap is likely to outperform LogMap on the FMA-NCI task as well. 
\bertmap can also significantly outperform the machine learning-based baseline $\text{LogMap-ML}^{*}$ on all the three tasks. This is because $\text{LogMap-ML}^{*}$ relies on LogMap and heuristic rules to extract high quality samples (anchor mappings) for training, but this strategy is not effective on our data. In contrast, \bertmap primarily relies on unsupervised data (synonyms and non-synonyms) to fine-tune the BERT model.


By comparing different \bertmap settings, we have the following observations. First, the semi-supervised setting ($+co$) is generally better than the unsupervised setting (without $co$), implying that \bertmap can effectively learn from given mappings. Second, complementary corpus is helpful especially when the task-involved ontologies are deficient in class labels --- on the FMA-SNOMED task, \bertmap with the complementary corpus ($+cp$) attains a higher F1 score than string-matching, edit-similarity, LogMapLt and LogMap-ML$^{*}$ baselines, all of which rely on class labels from within the input ontologies, by around $50\%$. Third, considering the identity synonyms ($+ids$) may slightly improve the performance or make no difference. Finally,  mapping extension and repair can consistently boost the performance, but not by much, possibly because it is hard to improve given that \bertmap's prediction part has already achieved high performance.

\begin{figure}[t!]
\begin{center}
    \includegraphics[width=0.45\textwidth]{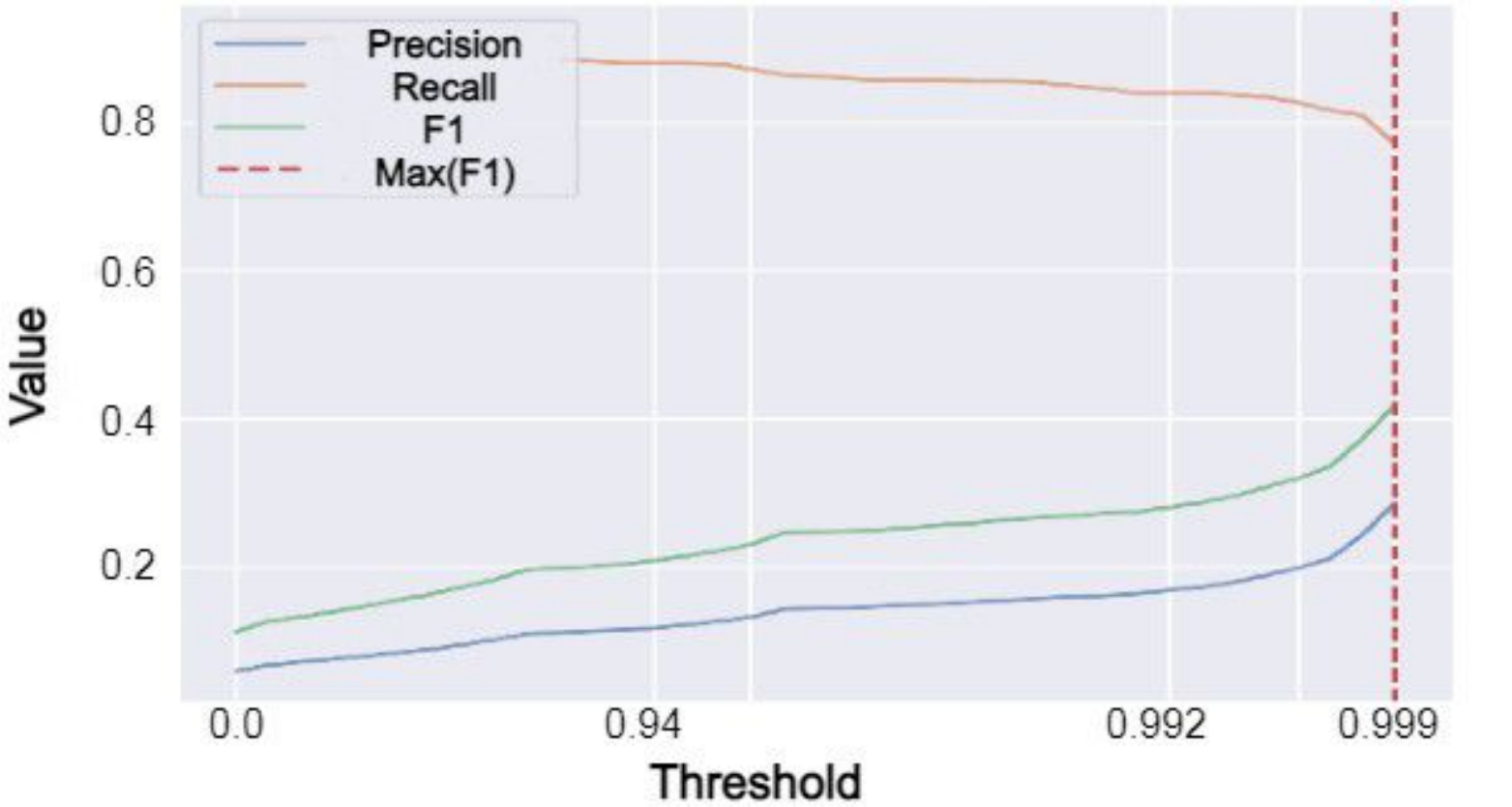}
\caption{ Validation results of \bertmap ($io+co+ids$) on the FMA-SNOMED+ task with mapping score threshold $\lambda$ ranging from $0$ to $1$.}
\label{fig:threshold}
\end{center}
\end{figure}

It is interesting to notice that \bertmap is robust to hyperparameter selection; most of its settings lead to the same best hyperparameters (i.e. $\tau = \texttt{src2tgt}$ and $\lambda=0.999\}$) on the validation set, $\mathcal{M}_{val}$. To further investigate this phenomenon, we visualize the validation process by presenting the plots of evaluation metrics against $\lambda$ in Figure \ref{fig:threshold}, where we can see that as $\lambda$ increases, Precision increases significantly while Recall drops only slightly --- thus F1 increases and attains the maximum at $\lambda = 0.999$. This observation is consistent for all \bertmap models in this paper\footnote{
See appendix for complete ablation study results.
}.

In Table \ref{tab:examples}, we present some examples of reference mappings that are retrieved by \bertmap but not by LogMap or AML. We can clearly see that, the BERT classifier captures the implicit connection between \textit{``third cervical''} and \textit{``C3''} in the first example, \textit{``posteior''} and \textit{``dorsal''} in the second example, as well as \textit{``wall''} and \textit{``membrane''} in the third example. This demonstrates the strength of contextual embeddings over the traditional lexical matching.

\begin{table}[t!]
    \centering
    \begin{adjustbox}{width=0.46\textwidth}
    \begin{tabular}{l l}
    \toprule
        FMA Class & SNOMED Class \\\midrule
        \multirow{2}{13em}{Third\_cervical\_spinal\_ganglion} & \multirow{2}{10.5em}{C3\_spinal\_ganglion} \\
        {} & {} \\
        \multirow{2}{13em}{Deep\_posterior\_sacrococcygeal ligament} & \multirow{2}{10.5em}{Structure\_of\_deep\_dorsal
        \_sacrococcygeal\_ligament} \\
        {} & {} \\ [0.1cm]
        \multirow{2}{13em}{Wall\_of\_smooth\_endoplasmic
        \_reticulum} & \multirow{2}{10.5em}{Agranular\_endoplasmic
        \_reticulum\_membrane} \\
        {} & {} \\
    \bottomrule
    \end{tabular}
    \end{adjustbox}
    \caption{
    Typical examples of reference mappings that are predicted by \bertmap but not by LogMap or AML.
    }
    \label{tab:examples}
\end{table}




\section{Related Work}

Classic OM systems are often based on lexical matching, structure matching and logical inference \cite{otero2015ontology}. For example, LogMap \cite{logmap} uses a lexical index to compute anchor mappings, then alternates between mapping extension that utilizes ontology structure, and mapping repair that utilizes logical reasoning; whereas AML \cite{amrlight} mixes several strategies to calculate lexical matching scores, followed by mapping extension and repair. Although these systems have proven quite effective, 
their lexical matching 
only utilizes the surface form of texts and ignores word semantics. \bertmap employs a similar architecture but utilizes BERT so that textual semantics and contexts
are considered in mapping computation.

Recent supervised learning-based OM approaches mainly focus on constructing class embeddings or extracting features. 
\citet{nkisi2018ontology} uses hand-crafted features such as string similarities together with Word2Vec; OntoEmma \cite{ontoalign-in-bio} relies on both hand-crafted features and word context features learned by a complex network; LogMap-ML \cite{chen2021augmenting} utilizes path contexts and ontology tailored word embeddings by OWL2Vec* \cite{chen2021owl2vec}; VeeAlign \cite{Iyer2020VeeAlignAS} proposes ``dual attention'' 
for class embeddings. However, these approaches often heavily depend on complicated feature engineering and/or complex neural networks. 
More importantly, they need a significant number of high quality labeled mappings for training which are often not available and costly to manually annotate.
Although some solutions such as distant supervision \cite{chen2021augmenting} and sample transfer \cite{nkisi2018ontology} were investigated, the sample quality varies and limits their performance. 
Unsupervised learning approaches such as ERSOM \cite{xiang2015ersom} and DeepAlign \cite{kolyvakis-etal-2018-deepalignment} were also studied. They attempt to refine word embeddings by, e.g., counter-fitting, to directly compute class similarity. However, they do not consider word contexts.

\citet{Neutel2021TowardsAO} have presented a preliminary OM investigation using BERT. Their work considered two relatively naive approaches: \textit{(i)} encoding classes with pre-trained BERT's token embeddings and calculating their cosine similarity; \textit{(ii)} fine-tuning class embeddings with the SentenceBERT \cite{Reimers2019SentenceBERTSE} 
architecture, which relies on a large number of given mappings. We have implemented \textit{(i)} and found it to perform much worse than string-matching on our tasks; moreover, according to their evaluation, method \textit{(ii)} has much lower mean reciprocal rank score than the non-contextual word embedding model, FastText \cite{Bojanowski2017EnrichingWV}, although it has higher coverage. Furthermore, their evaluation data have no gold standards, and thus, Precision, Recall and F1 are not computed.

\section{Conclusion and Future Work
}

In this paper, we propose a novel, general and practical OM system, \bertmap, which exploits the textual, structural and logical information of ontologies. The backbone of \bertmap is its predictor, which utilizes the contextual embedding model, BERT, to learn word semantics and contexts effectively, and computes mapping scores with the aid of sub-word inverted indices. The mapping extension and repair modules further improve the recall and precision, respectively.
\bertmap
works well with just the to-be-aligned ontologies and can be further improved by given mappings 
and/or complementary sources. 
%
In future, we will evaluate \bertmap with more large-scale (industrial) data.
We will also consider e.g., BERT-based ontology embedding for more robust mapping prediction, and more paradigms for integrating mapping prediction, extension and repair.

\section{Acknowledgments}
This work was supported by the SIRIUS Centre for Scalable Data Access (Research Council of Norway, project 237889), eBay, Samsung Research UK, Siemens AG, and the EPSRC projects OASIS (EP/S032347/1), UK FIRES (EP/S019111/1) and ConCur (EP/V050869/1).

\bibliography{mybib}

\appendix
\section{A \hspace{5pt} Full Ablation Results of Mapping Thresholds on the Validation Mapping Sets}

\begin{figure*}[t!]
\centering
\includegraphics[width=0.65\textwidth]{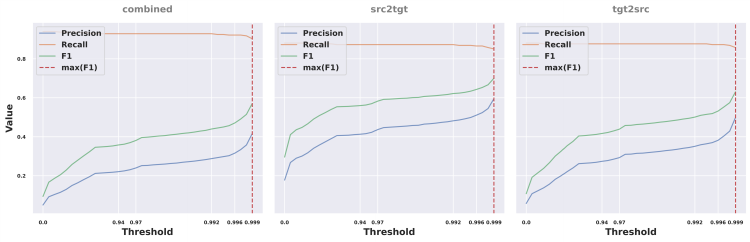}

\includegraphics[width=0.65\textwidth]{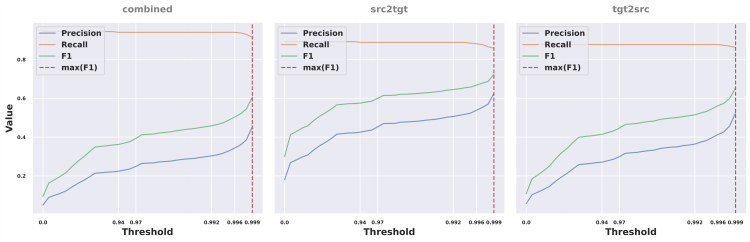}

\includegraphics[width=0.65\textwidth]{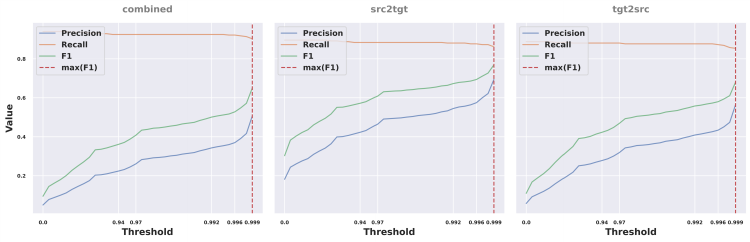}

\includegraphics[width=0.65\textwidth]{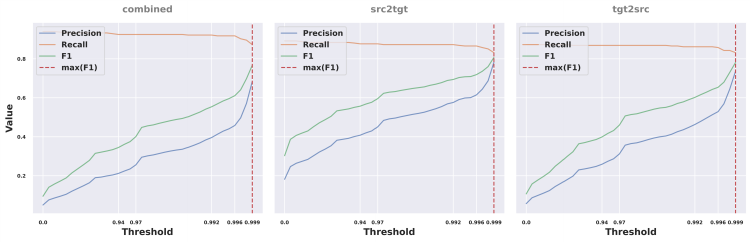}
\caption{\footnotesize  Precision, Recall and Macro-F1 of \bertmap on the validation sets of the FMA-NCI task as the mapping score threshold ranges from $0$ to $1$ (excluded $1$ because it represents the sting-match result). 
The maximum F1 is indicated by a red vertical line. }
\label{fig:fma2nci-val-thresholds}
\end{figure*}

\begin{figure*}[t!]
\centering

\includegraphics[width=0.6\textwidth]{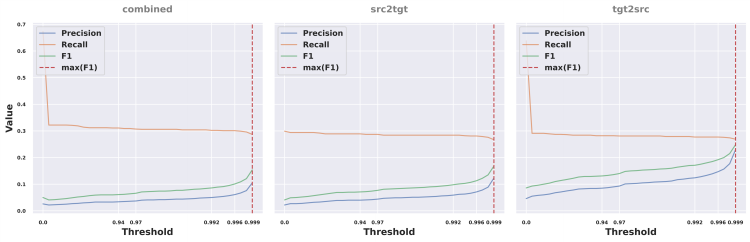}

\includegraphics[width=0.6\textwidth]{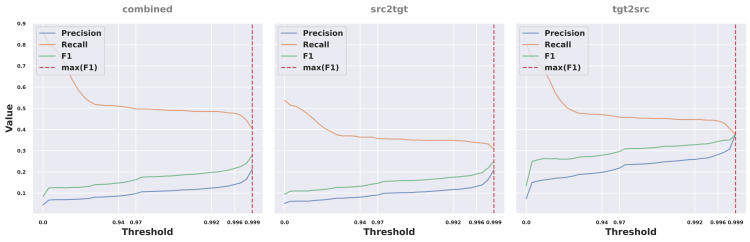}

\includegraphics[width=0.6\textwidth]{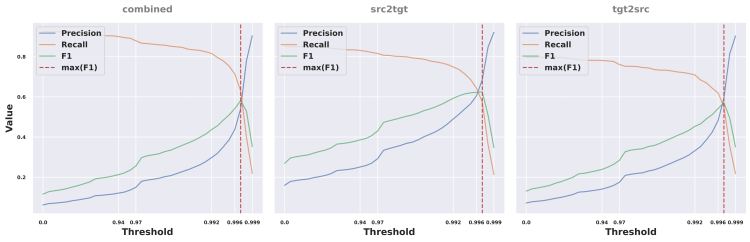}

\includegraphics[width=0.6\textwidth]{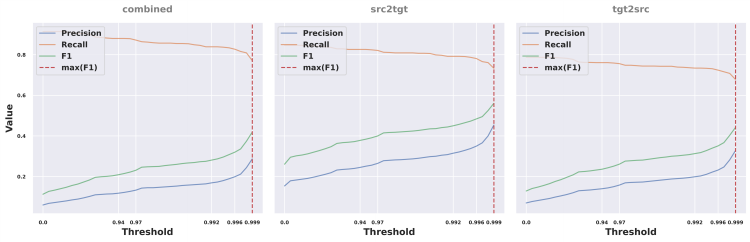}

\includegraphics[width=0.6\textwidth]{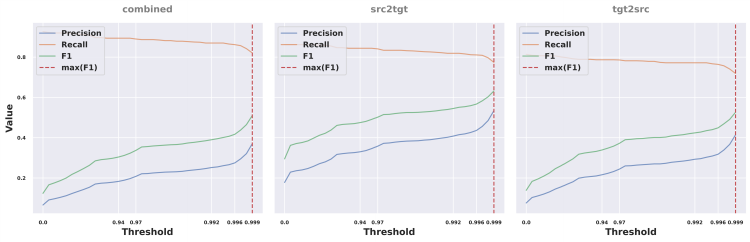}

\includegraphics[width=0.6\textwidth]{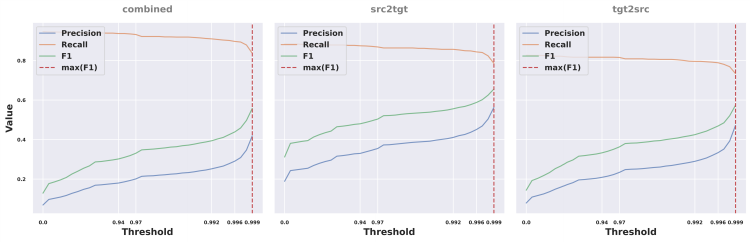}
\caption{\footnotesize  Precision, Recall and Macro-F1 of \bertmap on the validation sets of the FMA-SNOMED task as the mapping score threshold ranges from $0$ to $1$ (excluded $1$ because it represents the sting-match result). 
The maximum F1 is indicated by a red vertical line. }
\label{fig:fma2snomed-val-thresholds}
\end{figure*}

\begin{figure*}[t!]
\centering
\includegraphics[width=0.65\textwidth]{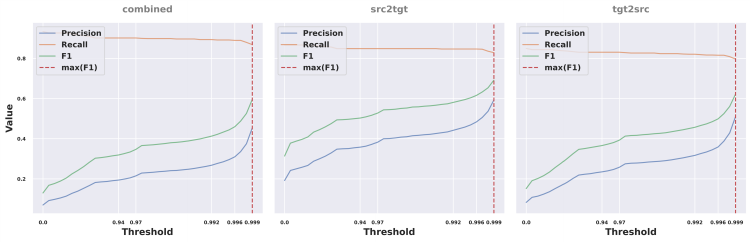}

\includegraphics[width=0.65\textwidth]{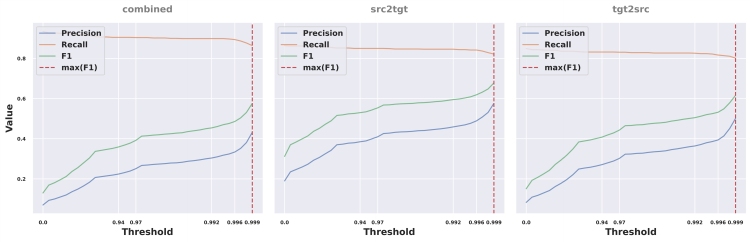}

\includegraphics[width=0.65\textwidth]{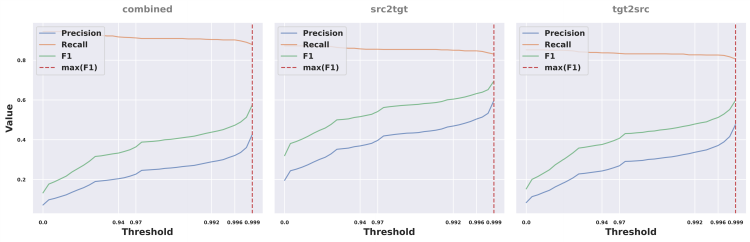}

\includegraphics[width=0.65\textwidth]{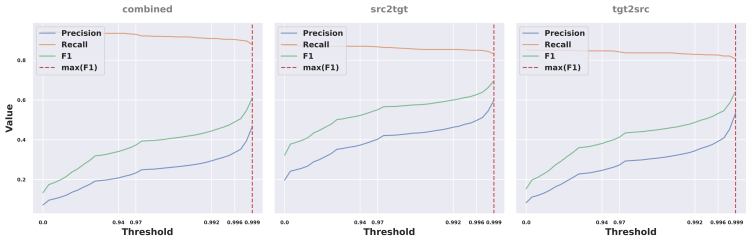}
\caption{\footnotesize  Precision, Recall and Macro-F1 of \bertmap on the validation sets of the FMA-SNOMED+ task as the mapping score threshold ranges from $0$ to $1$ (excluded $1$ because it represents the sting-match result). 
The maximum F1 is indicated by a red vertical line. }
\label{fig:fma2snomed+-val-thresholds}
\end{figure*}

In Figure \ref{fig:fma2nci-val-thresholds}, \ref{fig:fma2snomed-val-thresholds} and \ref{fig:fma2snomed+-val-thresholds}, we present, for all the \bertmap models in this paper, the plots of evaluation metrics (Precision, Recall and Macro-F1) against the mapping threshold $\lambda \in [0, 1)$ on the validation set. Figure \ref{fig:fma2snomed-val-thresholds} correspond to (left-to-right, top-to-bottom) the \texttt{combined}, \texttt{src2tgt}, and \texttt{tgt2src} results of $io$, $io+ids$, $io+co$, $io+co+ids$, $io+ids+cp$, $io+co+ids+cp$ settings on the FMA-SNOMED task. Figure \ref{fig:fma2nci-val-thresholds} and \ref{fig:fma2snomed+-val-thresholds} correspond to the \texttt{combined}, \texttt{src2tgt}, and \texttt{tgt2src} results of $io$, $io+ids$, $io+co$, $io+co+ids$ settings on the FMA-NCI task and FMA-SNOMED+ task, respectively.

Note that the validation results are generally worse than testing results because when evaluating on smaller mapping set, we need to ignore more positive mappings whereas the number of negative mappings stays the same, resulting in the prominent drop of F1 score.

\end{document}